\documentclass[final,authoryear,3p,times,12pt]{elsarticle}
\usepackage{lineno} 

\usepackage{graphics}
\usepackage{threeparttable} 
\usepackage{graphicx}
\usepackage{geometry}
\usepackage{multirow}
\usepackage{float} 
\usepackage{array}
\usepackage{makecell}
\usepackage{booktabs}
\usepackage{multirow}
\usepackage{makecell}
\usepackage{tabularx}
\usepackage{amsmath} 
\usepackage{indentfirst}
\usepackage{titlesec}
\usepackage{xcolor}
\usepackage{enumitem}
\titleclass{\subsubsubsection}{straight}[\subsection]
\newcounter{subsubsubsection}[subsubsection]
\renewcommand\thesubsubsubsection{\thesubsubsection.\arabic{subsubsubsection}.}
\titleformat{\subsubsubsection}
  {\normalfont\normalsize\itshape}{\thesubsubsubsection}{0.5em}{\indent}
\titlespacing*{\subsubsubsection}
  {0pt}{0pt}{0pt} 

\makeatletter
\renewcommand\paragraph{\@startsection{paragraph}{5}{\z@}%
  {3.25ex \@plus1ex \@minus.2ex}%
  {-1em}%
  {\normalfont\normalsize\itshape}}
\renewcommand\subparagraph{\@startsection{subparagraph}{6}{\parindent}%
  {3.25ex \@plus1ex \@minus .2ex}%
  {-1em}%
  {\normalfont\normalsize\itshape}}
\makeatother

\usepackage[hidelinks]{hyperref}
\definecolor{myblue}{RGB}{107,199,255}

\newcommand{\myfigref}[1]{\textcolor{myblue}{Fig.~\ref{#1}}}

\usepackage{bookmark} 
\usepackage{amssymb}
\usepackage{longtable}
\usepackage{ulem}

\usepackage{tikz}
\newcommand*{\circled}[1]{\lower.7ex\hbox{\tikz\draw (0pt, 0pt)%
    circle (.5em) node {\makebox[1em][c]{\small #1}};}}

\renewcommand{\eqref}[1]{\textup{\normalfont(\ref{#1})}}

\begin{document}
\begin{frontmatter}


\title{GSM-UTCI: A Multimodal Deep Learning Framework for High-Resolution Urban Thermal Comfort Prediction and Greening Scenario Analysis in Philadelphia}

\title{Planning for Cooler Cities: A Multimodal AI Framework for Predicting and Mitigating Urban Heat Stress through Urban Landscape Transformation}



\author[organization1]{Shengao Yi, \corref{cor}}
\author[organization1]{Xiaojiang Li}
\author[organization2]{Wei Tu}
\author[organization3]{Tianhong Zhao}

\address[organization1]{Department of City and Regional Planning, University of Pennsylvania, Philadelphia, PA 19104, USA}

\address[organization2]{Guangdong Key Laboratory for Urban Informatics, Guangdong-Hong Kong-Macao Joint Laboratory for Smart Cities, and Shenzhen Key Laboratory of Spatial Information Smart Sensing and Services, and Department of Urban Informatics, School of Architecture and Urban Planning, Shenzhen University, Shenzhen 518060, China}

\address[organization3]{College of Big Data and Internet, Shenzhen Technology University, Shenzhen, China}

\cortext[cor]{Corresponding author}

\vspace{30pt}
\begin{abstract}
As extreme heat events intensify due to climate change and urbanization, cities face increasing challenges in mitigating outdoor heat stress. While traditional physical models such as SOLWEIG and ENVI-met provide detailed assessments of human-perceived heat exposure, their computational demands limit scalability for city-wide planning. In this study, we propose GSM-UTCI, a multimodal deep learning framework designed to predict daytime average Universal Thermal Climate Index (UTCI) at 1-meter hyperlocal resolution. The model fuses surface morphology (nDSM), high-resolution land cover data, and hourly meteorological conditions using a feature-wise linear modulation (FiLM) architecture that dynamically conditions spatial features on atmospheric context. Trained on SOLWEIG-derived UTCI maps, GSM-UTCI achieves near-physical accuracy, with an $R^2$ of 0.9151 and MAE of 0.41 $^\circ$C, while reducing inference time from hours to under five minutes for an entire city. 
To demonstrate its planning relevance, we apply GSM-UTCI to simulate systematic landscape transformation scenarios in Philadelphia, replacing bare earth, grass, and impervious surfaces with tree canopy. Results show spatially heterogeneous but consistently strong cooling effects, with impervious-to-tree conversion producing the highest aggregated benefit (–4.18 $^\circ$C average $\Delta$UTCI across 270.7\,km$^2$). Tract-level bivariate analysis further reveals strong alignment between thermal reduction potential and land cover proportions. These findings underscore the utility of GSM-UTCI as a scalable, fine-grained decision support tool for urban climate adaptation, enabling scenario-based evaluation of greening strategies across diverse urban environments.
\end{abstract}

\end{frontmatter}

\section{Introduction}
\label{s1}

Cities around the world are experiencing increasingly severe and frequent heat stress due to the dual pressures of rapid urbanization and global climate change \citep{luo2018increasing, argueso2015effects, li2024sensitivity}. Urban areas often exhibit elevated temperatures compared to their rural surroundings, a phenomenon known as the urban heat island (UHI) effect \citep{mohajerani2017urban, deilami2018urban}, which exacerbates thermal discomfort \citep{lee2017overview}, raises energy demand \citep{li2019urban}, and intensifies health risks \citep{heaviside2017urban}, particularly for low-income and vulnerable populations \citep{chakraborty2019disproportionately, yuan2025surface}. The disproportionate exposure to heat across neighborhoods has brought urban heat mitigation to the forefront of planning, equity, and sustainability agendas \citep{keith2022planning, wilson2020urban}.

A key driver of urban heat lies in land surface characteristics: impervious materials such as asphalt and concrete absorb and retain heat, while vegetated surfaces like tree canopies mitigate heat through shading and evapotranspiration \citep{wang2019impacts, yi2025assessing, berry2013tree}. As such, the spatial composition of the urban landscape plays a fundamental role in shaping local microclimates \citep{zhou2011does, yang2023urban}. For example, \cite{li20233d} combined spatial gradient sampling method and multi scenarios simulations using the ENVI-met model to explore the reltionship between heat fluxes and microclimate in Beijing, China. They found that planting more trees in high sensible heat flux and low latent heat flux neighborhood can improve the heat environment. However, while this relationship is well established, urban planners often lack tools that can quantify and visualize how specific landscape changes might alter outdoor thermal conditions at the city scale.



    

Despite growing awareness of urban heat risks, the tools available to planners and researchers for modeling outdoor thermal comfort remain limited in scalability and practicality. Physics-based models such as SOLWEIG (Solar and Longwave Environmental Irradiance Geometry) and ENVI-met provide detailed simulations of radiative exchanges, shadowing, and energy balance, but their computational intensity makes them challenging to apply at the city scale \citep{lindberg2008solweig, bruse1998simulating, gal2020modeling}. As a result, their use is often constrained to small study areas or idealized urban forms \citep{yang2021verifying, salata2016urban}. 
In contrast, statistical and empirical approaches offer greater speed and flexibility but typically sacrifice spatial fidelity and generalizability. Many rely on coarse-resolution land surface temperature (LST) data or simplified assumptions about built form and vegetation, limiting their ability to inform hyperlocal interventions \citep{mao2021resolution, weng2014generating, feng2015enhancing}. Moreover, few existing studies systematically quantify how different land cover types, such as impervious surfaces, grass, or bare earth, individually and collectively influence heat stress across heterogeneous urban landscapes. 

Recent advances in artificial intelligence (AI) and geospatial data availability have opened new directions for modeling urban thermal environments. A growing number of studies have applied machine learning and deep learning techniques to estimate LST \citep{pande2024predictive, li2019evaluation}, thermal comfort indices \citep{brode2024application, zhong2022convolutional}, and related environmental variables \citep{subramaniam2022artificial, yi2025sub}. 
These data-driven approaches offer significant advantages in speed and scalability compared to traditional physical models.
However, most existing AI-based models are limited in three ways: they often predict coarse-scale LST rather than human-perceived heat stress metrics such as the Universal Thermal Climate Index (UTCI) and Mean Radiant Temperature ($T_{mrt}$) \citep{jendritzky2012utci, li2024sensitivity}; they rarely integrate multiple modalities of spatial and temporal data (e.g., surface morphology, land cover, and meteorology); and they typically do not support forward simulations of land-use or landscape change scenarios.

To address these gaps, we propose GSM-UTCI, a multimodal deep learning framework designed to predict daytime average UTCI at 1-meter resolution across entire urban areas. The model fuses three key data streams, surface morphology, land cover classification, and hourly meteorological conditions, through a Feature-wise Linear Modulation (FiLM) mechanism that dynamically conditions spatial features on atmospheric context. GSM-UTCI is trained on UTCI maps generated by SOLWEIG but achieves comparable accuracy while significantly improving computational efficiency: the model can generate 1-meter resolution UTCI predictions for an entire city in under five minutes, reducing runtime by orders of magnitude compared to traditional physical methods. Although developed primarily as a predictive model, GSM-UTCI also supports scenario-based simulations of land cover transformation, enabling climate-responsive planning and design interventions at actionable spatial scales.





    




\section{Literature review}
\label{rw}

\subsection{Urban heat stress and landscape structure}




Urban heat stress has emerged as a significant challenge for cities worldwide, driven by the combined effects of rapid urbanization and accelerating climate change \citep{argueso2015effects, he2023progress, luo2018increasing}. As global temperatures rise and urban populations grow denser, cities increasingly experience elevated thermal loads, increasing the risk of heat-related health impacts, particularly during summer periods \citep{klein2021vast, santamouris2020recent}. This intensification of urban heat exposure poses critical implications for public health \citep{singh2020urban, yang2024heat}, energy consumption \citep{santamouris2015impact, shahmohamadi2011impact}, and overall urban livability \citep{kashi2024effects, liang2020assessment}, making it a pressing concern for urban and landscape planners. More importantly, heat stress does not impact all urban residents equally. Vulnerable groups, including low-income populations, the elderly, and communities with limited access to green spaces are disproportionately exposed to higher temperatures and suffer greater adverse effects \citep{gronlund2016vulnerability, leap2024effects, chakraborty2019disproportionately}. For example, \cite{mitchell2015landscapes} compared the environmental justice results of heat risk in three largest US cities: New York City, Los Angeles, and Chicago. They found that there is a consistent and significant relationship between low-income community and minority status and higher urban heat risk. These disparities highlight the role of urban spatial structure and land management practices in mediating environmental risk.


Fundamental to the urban thermal environment are the surface characteristics of the landscape \citep{peng2016urban, li2020analysis, xie2020dominant}. Impervious surfaces such as roads, rooftops, and parking lots absorb and retain heat, increasing local temperatures \citep{chithra2015impacts, yun2011study, barnes2001impervious}, while vegetated areas, including tree canopies, grasslands, and wetlands, moderate microclimates through shading, evapotranspiration, and the alteration of surface radiation balance \citep{yi2025assessing, hesslerova2019wetlands, breshears1998effects}. Water bodies also contribute to local cooling effects via evaporation and thermal inertia \citep{wang2017surface}. Urban greenery, particularly tree canopy cover, plays a critical role in mitigating heat stress by intercepting solar radiation, reducing surface and air temperatures, and enhancing outdoor thermal comfort \citep{wong2021greenery, gillerot2024urban, cheela2021combating}. Research consistently demonstrates that areas with greater vegetation density exhibit significantly lower land surface temperatures compared to heavily built-up zones.

Recent studies have demonstrated the importance of three-dimensional landscape structure in shaping outdoor thermal environments. For example, \cite{kong2022impact} showed that metrics such as above-ground biomass, sky view factor, and building compactness significantly influence spatial patterns of mean radiant temperature, highlighting the cooling benefits of vegetation and the warming effects of compact urban forms. 
Overall, urban landscape structure, including the type, distribution, and connectivity of surface elements fundamentally shapes thermal conditions within cities. Through deliberate planning and landscape interventions, it is possible to strategically reconfigure urban form to reduce heat exposure, improve thermal equity, and enhance the resilience of cities to climate-related stresses.

\subsection{Traditional methods for heat stress modeling}


Traditional approaches for modeling outdoor thermal comfort and UTCI conditions have primarily relied on physics-based simulations \citep{li2024sensitivity}. Models such as SOLWEIG and ENVI-met have been widely used to simulate complex urban microclimates by accounting for radiation fluxes, surface energy balances, air temperature, humidity, and wind fields at fine spatial and temporal resolutions \citep{lindberg2008solweig, bruse1998simulating, gal2020modeling}. These models provide valuable insights into the localized impacts of urban morphology, vegetation, and built structures on human thermal exposure \citep{badino2021modelling, li2023comparing, hosseinihaghighi2020using}.


However, despite their detailed physical foundations, traditional modeling approaches present significant limitations when applied to large-scale urban environments. One major constraint is the high computational cost associated with simulating detailed energy balances across extensive urban areas at high spatial resolution. Even efforts to accelerate the modeling process, such as the GPU-based optimization of SOLWEIG proposed by \cite{li2021gpu}, have only partially addressed this challenge; depending on model complexity and data size, simulating UTCI for an entire city can still require processing times ranging from several tens of minutes to multiple hours. These computational demands, combined with the need for extensive input preparation and calibration, make traditional methods operationally challenging for planners and policymakers seeking rapid or iterative scenario evaluations.


In response to these challenges, statistical and empirical models have been proposed as faster alternatives for estimating urban heat exposure. For example, \cite{alkhaled2024webmrt} have developed WebMRT, an online platform utilizing machine learning algorithms such as LightGBM to rapidly estimate $T_{mrt}$ based on easily obtainable environmental and meteorological parameters. Similarly, \cite{brode2024application} evaluated the application of various statistical learning algorithms, including random forests and k-nearest neighbors, in predicting UTCI equivalent temperatures and associated thermal stress categories. Their findings indicated that while statistical learning approaches could achieve reasonable predictive accuracy (e.g., RMSE $\approx$ 3°C), clustering-based methods showed limited agreement with expert-defined thermal stress classifications. 
While these approaches demonstrate the potential to streamline thermal stress modeling, they often rely on simplified predictors and may struggle to capture the complex spatial heterogeneity inherent in urban environments. Consequently, many statistical models lack the spatial detail and generalizability needed for neighborhood-level planning and scenario-based landscape interventions.

\subsection{Data-driven methods for heat stress modeling}


Recent advances in AI have led to the emergence of machine learning (ML) and deep learning (DL) techniques as promising tools for modeling urban climate processes, including the prediction of heat stress. Unlike traditional physics-based models, which rely on solving radiative and thermodynamic equations, AI-driven approaches leverage multiple geospatial datasets, including remote sensing imagery, meteorological observations, and built environment features to learn complex and non-linear relationships that influence urban thermal environments. These methods offer substantial gains in computational efficiency and scalability, making them increasingly attractive for large scale assessments and real-time planning applications. 

A growing body of work has applied DL models to estimate $T_{mrt}$, UCTI and other heat-related metrics. For example, \cite{zhong2022convolutional} utilized convolutional neural networks (CNNs) to directly generate UTCI microclimate maps from spatial inputs, achieving results comparable to physical models with significantly faster processing times. \cite{xie2022prediction} combined multilayer neural networks with optimization algorithms to simulate $T_{mrt}$ distributions around building geometries, demonstrating high accuracy and practical feasibility for architectural-scale applications. At the global scale, \cite{yang2024gloutci} developed 1 km resolution UTCI datasets by integrating Sentinel satellite imagery with deep learning, advancing data availability for macro-scale climate resilience planning. In addition, \cite{briegel2024high} validated the utility of AI-based models for simulating urban thermal conditions at neighborhood scales, illustrating their potential to bridge human biometeorology with urban design.






Despite these advances, most existing AI-driven applications have focused on predicting instantaneous thermal conditions at specific time points, often representing peak afternoon hours. Few models have been designed to capture the diurnal variation of human-perceived heat stress, particularly by estimating average daytime UTCI. Moreover, although considerable progress has been made in fine-scale thermal environment mapping, relatively few studies systematically evaluate the thermal impacts of different urban landscape components (e.g., tree canopy, impervious surfaces, bare soil) through scenario-based simulation. As cities increasingly seek data-informed strategies for climate resilience, there remains a significant need for high-resolution, transferable modeling frameworks that can both predict spatial patterns of heat stress and simulate the potential effects of landscape transformation interventions to guide planning and design decisions.

\subsection{Landscape-based planning strategies for urban heat mitigation}


Urban and landscape planning strategies have increasingly recognized the role of land surface interventions in mitigating heat exposure \citep{semenzato2023urban, norton2015planning, lindberg2016impact, chen2022integrated}. Approaches such as expanding urban forestry, enhancing green space connectivity, and incorporating permeable surface materials have been widely promoted to improve microclimatic conditions and reduce urban heat stress \citep{pereira2024guidelines, bosch2021evaluating}. Among these, increasing tree canopy cover has consistently emerged as one of the most effective strategies for lowering surface and air temperatures, improving outdoor thermal comfort, and enhancing urban resilience to climate extremes \citep{kim2024greater}.



However, many existing planning recommendations are derived primarily from empirical observations, small-scale experimental studies, or localized field measurements \citep{yin2024cooling, middel2015urban}. While these studies provide important insights, they often lack the spatial breadth and predictive capacity needed to support city-wide intervention planning. Comprehensive, simulation-based evaluations that systematically estimate the cooling potential of different land cover transformation strategies across diverse urban contexts remain relatively rare \citep{schrodi2023climate}.



This gap poses a significant challenge for planners and designers who must make landscape intervention decisions at varying spatial scales and under diverse urban morphological conditions. Without robust, spatially explicit predictive tools, it is difficult to prioritize interventions, assess their cumulative impacts, or optimize urban greening efforts for maximum thermal benefit. 
Therefore, there is a pressing need for simulation-based approaches that can quantitatively assess the thermal impacts of landscape transformations at hyperlocal resolution. Such frameworks are critical for informing effective, equitable, and climate-resilient urban planning and landscape design interventions.

\section{Methodology}
\label{method}

\begin{figure}[H]
    \centering
    \includegraphics[width=\linewidth]{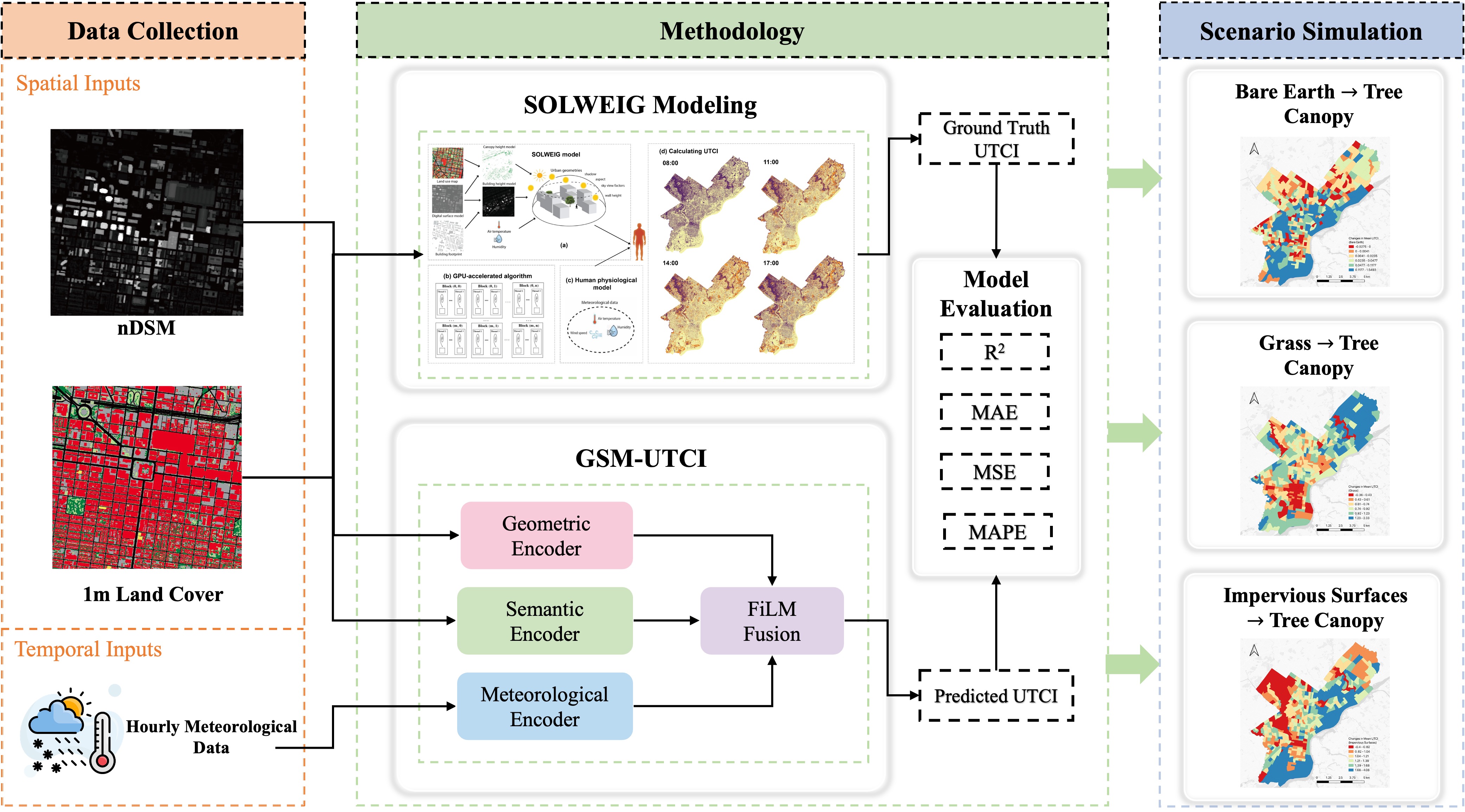}
    \caption{Overview of the proposed multimodal framework for high-resolution UTCI prediction and simulation. The framework integrates spatial and temporal data inputs, including 1-meter resolution normalized DSM (nDSM), land cover maps, and hourly meteorological data. SOLWEIG is first used to generate hourly ground truth UTCI maps across the city, serving as training labels. The GSM-UTCI model employs three specialized encoders (geometric, semantic, and meteorological) to extract features from spatial and temporal modalities. These representations are fused via a FiLM-based module that conditions spatial features on dynamic meteorological states. The model is evaluated and applied to scenario-based simulations of landscape transformations to assess their cooling impact.}
    \label{fig:framework}
\end{figure}

To predict high-resolution urban heat stress across complex cityscapes, we propose a multimodal deep learning framework that fuses spatial and temporal data sources. As shown in \myfigref{fig:framework}, the modeling pipeline begins with the collection of key spatial inputs, 1-meter normalized Digital Surface Models (nDSM) and land cover maps, alongside hourly meteorological variables. These inputs are used to drive the SOLWEIG model, which generates hourly UTCI maps that serve as the training data. The core predictive architecture, namely GSM-UTCI, consists of three parallel encoders: a geometric encoder for urban morphology, a semantic encoder for land surface properties, and a meteorological encoder that processes dynamic weather conditions. These heterogeneous representations are fused through a FiLM mechanism, where temporal features condition the spatial encodings. The predicted UTCI maps are evaluated against SOLWEIG outputs using multiple statistical metrics. Finally, the trained model supports city-scale simulation of land cover transformation scenarios, enabling planners to assess the thermal benefits of targeted interventions such as increasing tree canopy over impervious or bare surfaces.

\subsection{Study area}

\begin{figure}[H]
    \centering
    \includegraphics[width=0.7\linewidth]{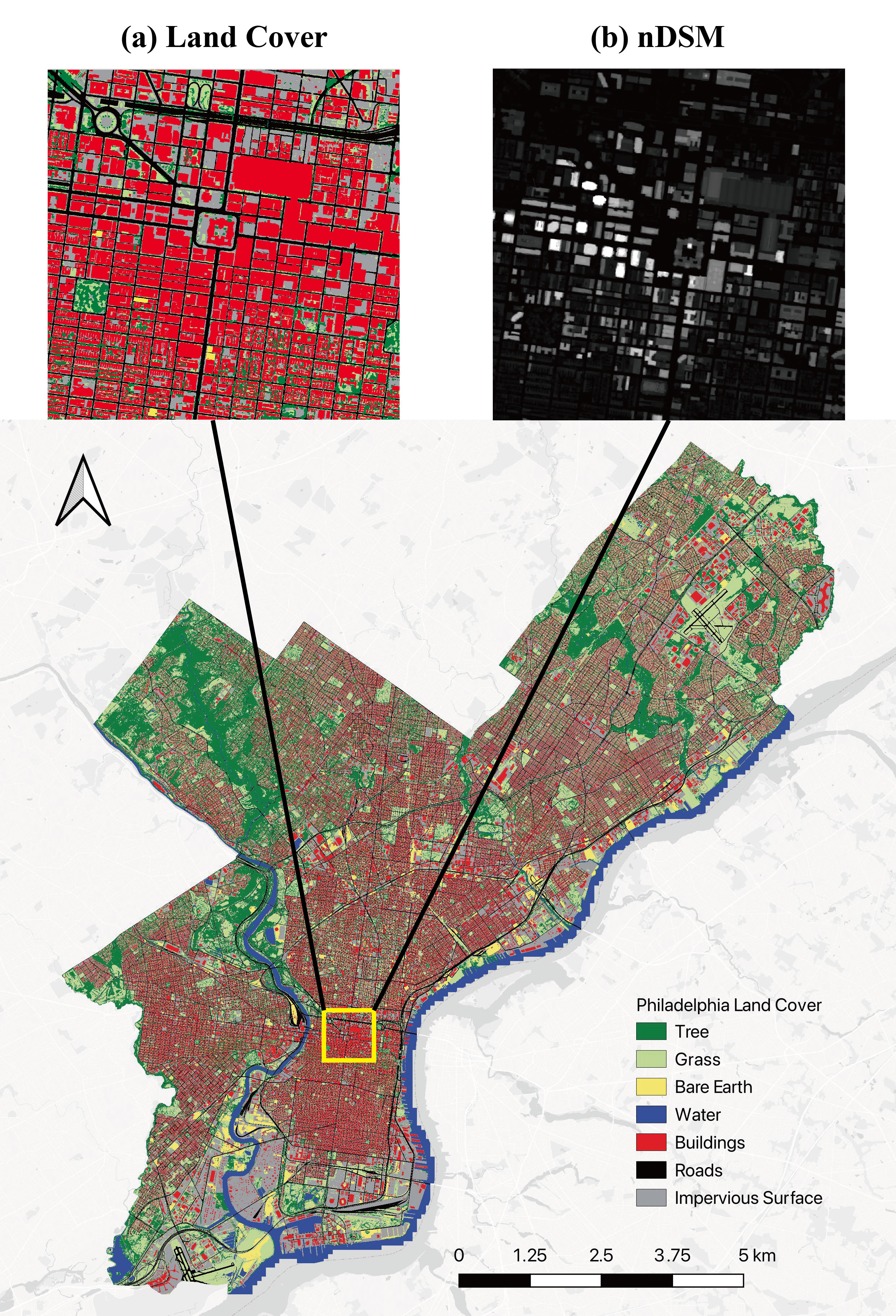}
\caption{The study area is Philadelphia, United States. (a) A patch of the land cover map in the study area, (b) the nDSM of a portion of the study area.}
    \label{fig_SA}
\end{figure}

The study area is Philadelphia, the sixth-most populous city in the United States, which is located in the southeastern region of Pennsylvania along the Delaware and Schuylkill Rivers. It experiences a humid subtropical climate, characterized by hot, humid summers and relatively mild winters, which intensifies concerns about urban heat exposure during peak summer months. As shown in \myfigref{fig_SA}, the city is made up with a diverse urban environments that includes dense downtown cores, low-rise residential neighborhoods, large park systems, industrial zones, and waterfront areas, making it an ideal area for analyzing intra-urban thermal variability. With a legacy of redlining, uneven green infrastructure distribution, and severe socioeconomic disparities, Philadelphia also presents critical challenges and opportunities for equitable climate adaptation. This study focuses on capturing the spatial heterogeneity of average summer UTCI across the entire city, leveraging high-resolution geospatial data to inform both technical model validation and policy-relevant greening interventions.

\subsection{Data sources}

This study integrates a range of high-resolution spatial and meteorological datasets, all corresponding to the year 2020 or close to it, to support UTCI prediction and scenario simulation. The 1-meter land use map, developed semi-automatically using high-resolution aerial imagery and LiDAR data, was obtained from the Pennsylvania Spatial Data Access (PASDA) (\href{https://www.pasda.psu.edu/}{https://www.pasda.psu.edu/}). This dataset includes detailed classifications such as tree canopy, grass, bare earth, water, buildings, roads, and impervious surfaces, and achieves an overall classification accuracy of approximately 90\%.

LiDAR point cloud data, in the form of pre-processed x, y, and z coordinate files, was sourced from the United States Geological Survey (USGS) 3D Elevation Program (\href{https://usgs.entwine.io/}{https://usgs.entwine.io/}. Using the open-source PDAL library, the point cloud data were processed into a Digital Elevation Model (DEM) and a Digital Surface Model (DSM). These elevation products were further used, along with the land use map and building footprint data, to generate high-resolution building height and tree canopy height models across the study area. Building footprint data with associated height attributes were collected from the City of Philadelphia’s Open Data Portal (\href{https://opendataphilly.org/datasets/building-footprints/}{https://opendataphilly.org/datasets/building-footprints/}). 

Hourly meteorological data were acquired from the National Solar Radiation Database (NSRDB), maintained by the National Renewable Energy Laboratory (NREL) (\href{https://nsrdb.nrel.gov/}{https://nsrdb.nrel.gov/}. The dataset includes 18 key atmospheric variables such as air temperature, relative humidity, global horizontal irradiance (GHI), direct normal irradiance (DNI), diffuse horizontal irradiance (DHI), and so on. This study focused the August, representing typical summer conditions in Philadelphia. These parameters form the temporal inputs to the model and support the calculation of the UTCI.

\subsection{UTCI modeling through SOLWEIG}

\begin{figure}[H]
    \centering
    \includegraphics[width=\linewidth]{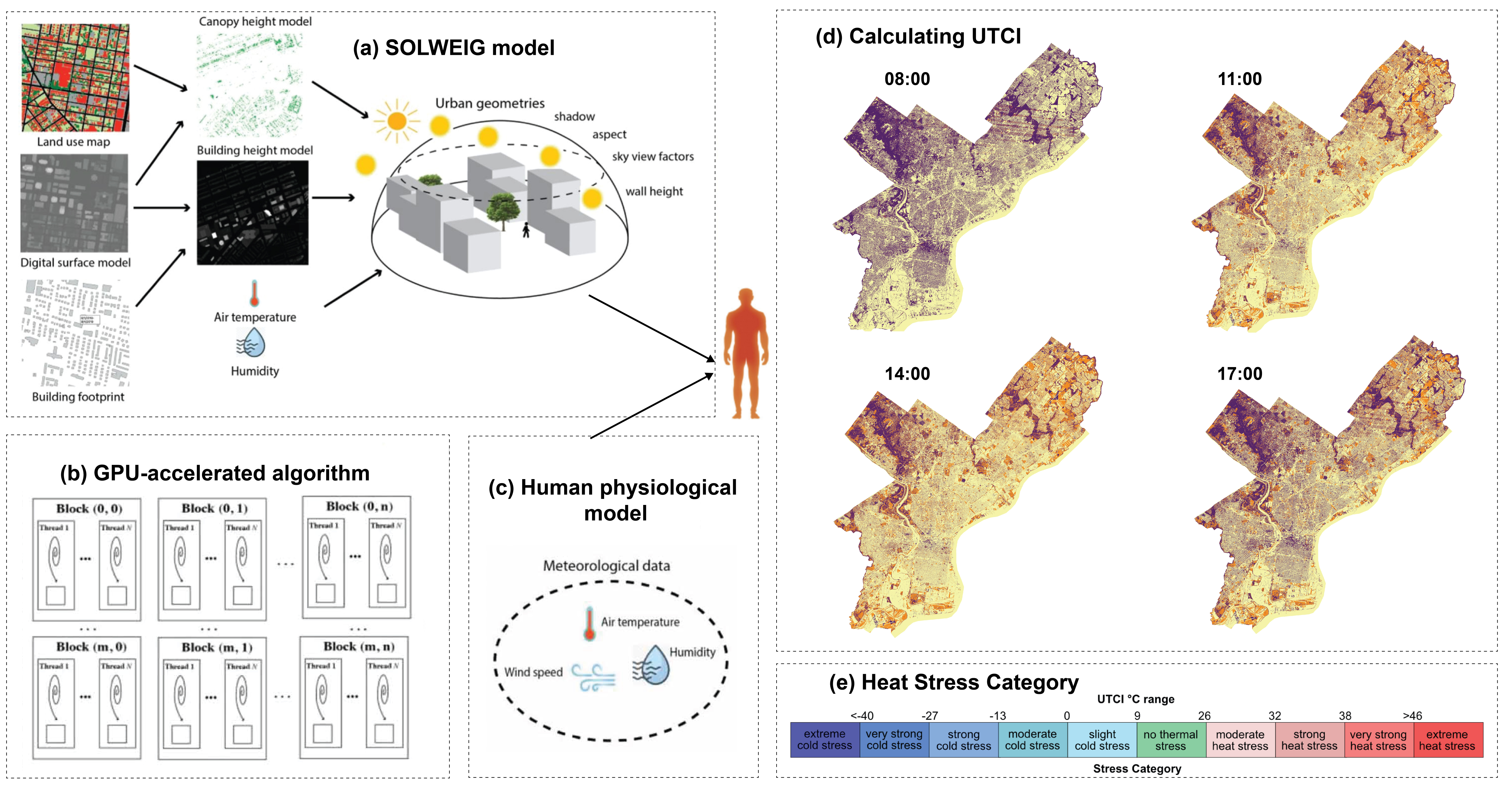}
\caption{The calculation of the UTCI and the $T_{mrt}$ combines the SOLWEIG and human physiological model based on tree canopy height model, building height model, and meteorological data, accelerated by GPU: a) the SOLWEIG model for calculation of the mean radiant temperature, b) the GPU-accelerated algorithm, c) human physiological model, d) spatio-temporal UTCI calculation, and e) category of heat stress.}
    \label{fig_UTCI}
\end{figure}

This study employed the UTCI to quantify human thermal stress in outdoor urban environments. The UTCI is a comprehensive indicator that accounts for the combined effects of air temperature, humidity, wind speed, and $T_{mrt}$, making it particularly suitable for assessing heat stress across complex urban landscapes. As shown in Figure~\ref{fig_UTCI}, UTCI values are classified into stress categories, with 32 $^\circ$C commonly used as the threshold for strong heat stress \citep{walikewitz2018assessment, li2024sensitivity}.

Among the input parameters, $T_{mrt}$ plays a pivotal role in determining thermal comfort, as it represents the net radiant energy absorbed by the human body from surrounding surfaces and the atmosphere. To estimate $T_{mrt}$, we employed the SOLWEIG model \citep{lindberg2008solweig}, a 3D radiative transfer model that simulates both shortwave and longwave radiation exchanges. The model considers urban geometry, surface orientation, shading, and view factors, making it well-suited for complex built environments.

Inputs to the SOLWEIG model included a high-resolution land use map, building height model, canopy height model, and hourly meteorological data (air temperature, humidity, and radiation components). The model calculates mean radiant flux ($R_{str}$) based on radiation in six directions—north, south, east, west, top, and bottom—using the following equation:

\begin{equation}
R_{str} = \zeta_k \sum_{i=1}^{6} K_i F_i + \varepsilon_p \sum_{i=1}^{6} L_i F_i
\label{eq_R_str}
\end{equation}

where $K_i$ and $L_i$ denote directional shortwave and longwave radiation fluxes, and $F_i$ are angular view factors. The absorption coefficient for shortwave radiation ($\zeta_k$) was set to 0.70, and the emissivity of the human body ($\varepsilon_p$) was set to 0.97. $T_{mrt}$ was then derived from $R_{str}$ using the Stefan–Boltzmann law:

\begin{equation}
T_{mrt} = \sqrt[4]{\frac{R_{str}}{(\varepsilon_p \sigma)}} - 273.15
\label{eq_T_mrt}
\end{equation}

where $\sigma$ is the Stefan–Boltzmann constant ($5.67 \times 10^{-8} , Wm^{-2}K^{-4}$). To efficiently handle high-resolution and city-wide computations, this study employed a previously developed GPU-accelerated version of the SOLWEIG model \citep{li2021gpu}, significantly reducing the runtime for large-scale radiative simulations.

Following the $T_{mrt}$ estimation, this study applied the official UTCI approximation algorithm \citep{brode2012deriving}, originally written in Fortran, and adapted it into a GPU-accelerated pipeline. UTCI was calculated hourly from 8:00 a.m. to 7:00 p.m. throughout August 2020. These hourly estimates were then averaged to generate a spatially continuous representation of typical summer UTCI conditions across Philadelphia. 
This high-resolution UTCI map serves both as a baseline reference and as a validation target for training the proposed GSM-UTCI deep learning framework.

\subsection{GSM-UTCI model architecture}

\begin{figure}[H]
    \centering
    \includegraphics[width=\linewidth]{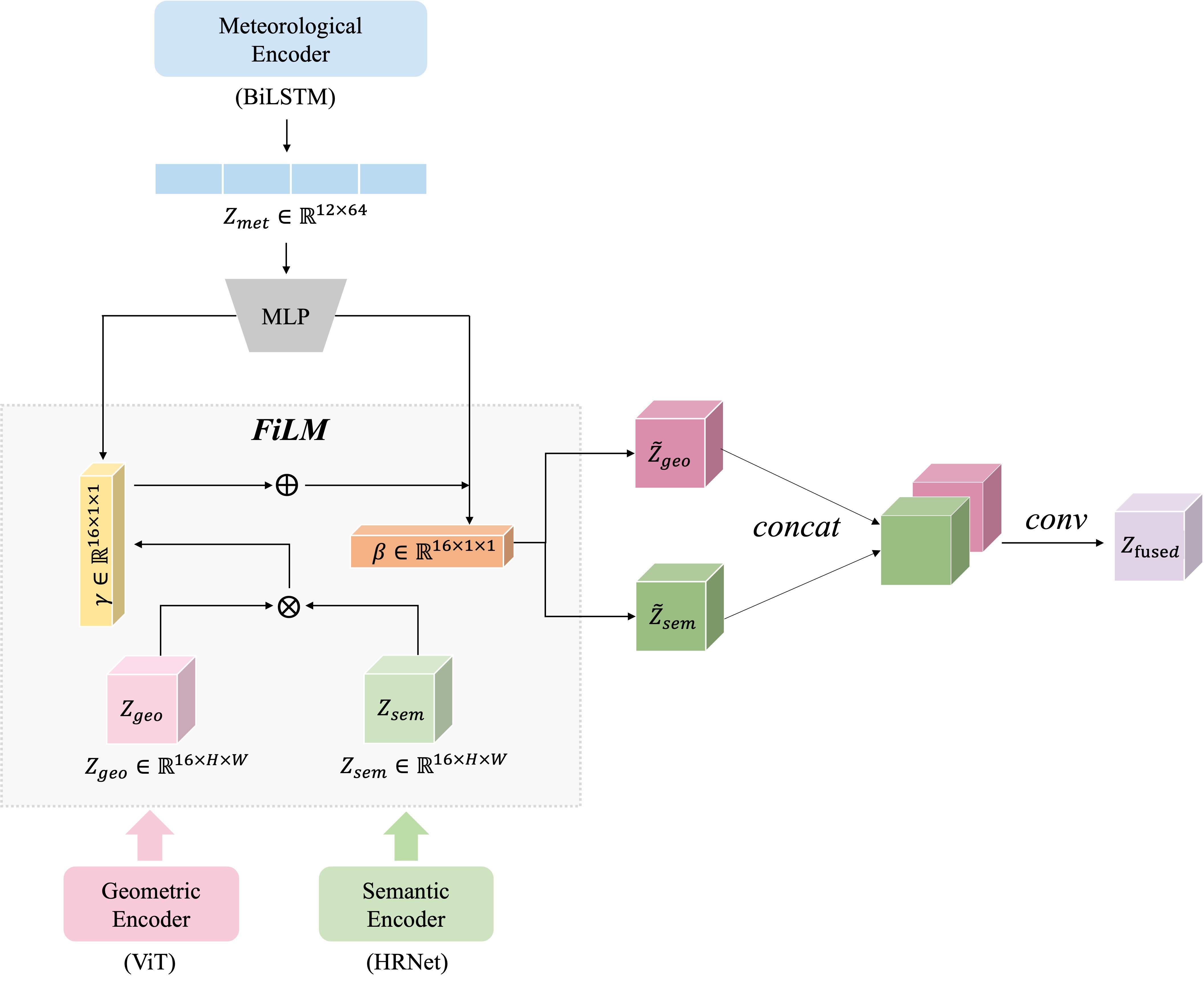}
    \caption{The architecture of GSM-UTCI.}
    \label{fig:GSM}
\end{figure}

The proposed model architecture, GSM-UTCI, is designed to predict spatially detailed daytime average UTCI maps by integrating three complementary streams of information: surface morphology, land cover, and meteorological dynamics. The architecture comprises three primary encoder modules. The geometric encoder extracts structural features from nDSM, providing a compact representation of urban form. The semantic encoder processes high-resolution land cover maps using a high-fidelity convolutional backbone to capture material and vegetative variation across the urban surface. The meteorological encoder leverages a BiLSTM architecture to model hourly sequences of weather and solar radiation data, producing temporally structured embeddings that represent cumulative thermal forcing throughout the day.

These three encoded modalities are fused through a Feature-wise Linear Modulation (FiLM) mechanism, wherein the meteorological features dynamically condition the spatial encoders via channel-wise scaling and shifting operations. The geometric and semantic spatial features, after FiLM modulation, are concatenated and processed through convolutional layers to produce the final UTCI prediction at 1-meter resolution. This multimodal design enables the GSM-UTCI framework to capture complex interactions between urban landscape structure and temporally evolving climatic conditions, supporting scenario-based simulation and high-resolution thermal comfort analysis across diverse urban environments.

\subsubsection{Geometric encoder}

To extract structural information from the urban surface, we design a geometric encoder based on a Vision Transformer (ViT) backbone. The input to this encoder is a nDSM, denoted as $\mathbf{X}_{\text{dsm}} \in \mathbb{R}^{1 \times H \times W}$, where $H$ and $W$ represent the spatial dimensions of the input tile at 1-meter resolution. 

To accommodate the ViT architecture, a stem convolution is first applied to transform the single-channel input into a 3-channel tensor compatible with pretrained weights:
\begin{equation}
\tilde{\mathbf{X}}_{\text{dsm}} = f_{\text{stem}}(\mathbf{X}_{\text{dsm}}), \quad \tilde{\mathbf{X}}_{\text{dsm}} \in \mathbb{R}^{3 \times H \times W}
\end{equation}

The ViT encoder, denoted as $f_{\text{ViT}}(\cdot)$, partitions the input into non-overlapping patches and models long-range dependencies across spatial locations using multi-head self-attention. The resulting token sequence is reshaped into a coarse spatial feature map:
\begin{equation}
\mathbf{F}_{\text{geo}} = f_{\text{ViT}}(\tilde{\mathbf{X}}_{\text{dsm}}) \in \mathbb{R}^{C' \times H' \times W'}
\end{equation}
where $C'$ is the intermediate embedding dimension, and $H', W'$ are determined by the ViT patch size (e.g., $H'=H/P$, $W'=W/P$ for patch size $P$).

A projection layer then reduces the channel dimension to $C=16$ via a $1 \times 1$ convolution:
\begin{equation}
\hat{\mathbf{F}}_{\text{geo}} = \text{Conv}_{1\times1}(\mathbf{F}_{\text{geo}}) \in \mathbb{R}^{16 \times H' \times W'}
\end{equation}

Finally, the feature map is upsampled back to the original resolution using bilinear interpolation:
\begin{equation}
\mathbf{Z}_{\text{geo}} = \text{Upsample}(\hat{\mathbf{F}}_{\text{geo}}) \in \mathbb{R}^{16 \times H \times W}
\end{equation}

This output $\mathbf{Z}_{\text{geo}}$ is used as the geometric feature representation in the GSM-UTCI model, capturing both vertical structure and contextual spatial patterns of the built environment.


\subsubsection{Semantic encoder}

To capture the material and surface-type characteristics of the urban landscape, we implement a semantic encoder based on a High-Resolution Network (HRNet) backbone. This module takes as input a high-resolution categorical land cover map, denoted as $\mathbf{X}_{\text{lc}} \in \mathbb{R}^{1 \times H \times W}$, where $H$ and $W$ represent the spatial dimensions of a tile at 1-meter resolution.

Since pretrained HRNet weights are typically optimized for 3-channel RGB images, we apply a $3 \times 3$ convolutional stem to map the single-channel input to a 3-channel tensor:
\begin{equation}
\tilde{\mathbf{X}}_{\text{lc}} = f_{\text{stem}}(\mathbf{X}_{\text{lc}}), \quad \tilde{\mathbf{X}}_{\text{lc}} \in \mathbb{R}^{3 \times H \times W}
\end{equation}

HRNet processes this input through parallel multi-resolution pathways and performs repeated feature exchange across different scales. Let $\{ \mathbf{F}_i \}_{i=1}^{L}$ denote the feature maps extracted at $L$ resolution levels, where each $\mathbf{F}_i \in \mathbb{R}^{C_i \times H_i \times W_i}$. These feature maps are individually upsampled to a common intermediate resolution (e.g., $H/4 \times W/4$) and then concatenated:
\begin{equation}
\mathbf{F}_{\text{concat}} = \text{Concat}\left( \text{Upsample}(\mathbf{F}_1), \ldots, \text{Upsample}(\mathbf{F}_L) \right)
\end{equation}

A $1 \times 1$ convolution is then applied to reduce the aggregated channels to a fixed output dimension $C = 16$, followed by bilinear upsampling to recover the original resolution:
\begin{equation}
\mathbf{Z}_{\text{sem}} = \text{Upsample}\left( \text{Conv}_{1\times1}(\mathbf{F}_{\text{concat}}) \right) \in \mathbb{R}^{16 \times H \times W}
\end{equation}

The output $\mathbf{Z}_{\text{sem}}$ encodes the spatial heterogeneity of the urban surface, capturing local variations in vegetation, buildings, impervious surfaces, and bare ground. These features are crucial for modeling the differential heating patterns that contribute to spatial variations in heat stress.


\subsubsection{Meteorological encoder}

To capture the temporal dynamics of meteorological and solar conditions throughout the day, we implement a meteorological encoder based on a bidirectional Long Short-Term Memory (BiLSTM) architecture. The input to this module is a multivariate time series, denoted as $\mathbf{X}_{\text{met}} \in \mathbb{R}^{T \times N}$, where $T = 12$ represents the number of hourly time steps (from 8 a.m. to 7 p.m.), and $N = 18$ denotes the number of meteorological and solar-related variables at each hour.

The BiLSTM module processes the sequence bidirectionally, enabling the encoder to capture both past and future dependencies. This structure allows the model to learn cumulative and lagged thermal effects, such as the interplay between solar radiation, humidity, temperature, and wind speed, that are critical for simulating realistic thermal stress.

Each time step is encoded into a latent embedding of dimension $d = 64$. Let $f_{\text{BiLSTM}}(\cdot)$ denote the encoding function, then the output of this module is a temporal feature matrix:
\begin{equation}
\mathbf{Z}_{\text{met}} = f_{\text{BiLSTM}}(\mathbf{X}_{\text{met}}) \in \mathbb{R}^{T \times d}
\end{equation}

Compared to simpler approaches such as hourly averaging, the BiLSTM-based representation provides richer temporal context and enhances the model's generalizability across diverse weather conditions. These temporal features are later fused with spatial encodings to generate high-resolution UTCI predictions.


\subsubsection{FiLM-based feature fusion module}


To integrate spatial and temporal representations in a context-aware manner, we design a fusion module based on Feature-wise Linear Modulation (FiLM). This approach enables the model to dynamically condition the influence of spatial features using meteorological context, thereby enhancing cross-modal interactions relevant to urban heat stress prediction.

In our framework, the geometric encoder produces a feature map $\mathbf{Z}_{\text{geo}} \in \mathbb{R}^{C \times H \times W}$ and the semantic encoder outputs $\mathbf{Z}_{\text{sem}} \in \mathbb{R}^{C \times H \times W}$. Simultaneously, the meteorological encoder yields a conditioning vector $\mathbf{Z}_{\text{met}} \in \mathbb{R}^{T \times d}$, summarizing the temporal variation in weather and solar-related variables throughout the day.

This vector is used as input to two parameter generation networks that produce FiLM parameters: a channel-wise scaling vector $\gamma$ and a shifting vector $\beta$ for each spatial encoder. Given an input feature map $\mathbf{Z}$, the FiLM modulation is defined as:
\begin{equation}
\tilde{\mathbf{Z}} = \gamma \cdot \mathbf{Z} + \beta
\end{equation}
where $\gamma, \beta \in \mathbb{R}^{C \times 1 \times 1}$ are broadcasted across the spatial dimensions and applied independently to each channel.

The modulated spatial features $\tilde{\mathbf{Z}}_{\text{geo}}$ and $\tilde{\mathbf{Z}}_{\text{sem}}$ are concatenated along the channel axis and fused via a sequence of convolutional layers:
\begin{equation}
\mathbf{Z}_{\text{fused}} = \text{Conv} \left( \text{Concat}(\tilde{\mathbf{Z}}_{\text{geo}}, \tilde{\mathbf{Z}}_{\text{sem}}) \right)
\end{equation}

The resulting fused representation integrates both spatial heterogeneity and temporal context. A final prediction head outputs the high-resolution UTCI map at 1-meter resolution:
\begin{equation}
\hat{\mathbf{Y}}_{\text{UTCI}} \in \mathbb{R}^{1 \times H \times W}
\end{equation}

By allowing meteorological information to modulate spatial encodings in a fine-grained and learnable way, the FiLM-based fusion mechanism improves the model’s ability to simulate urban heat stress under varying environmental conditions.

\subsection{Ablation and comparative studies}

To evaluate the effectiveness of our proposed GSM-UTCI model architecture, we design a set of ablation and comparison experiments using different encoder combinations and fusion strategies. All model variants are trained and evaluated under identical conditions using the same dataset split, tile size, and training schedule. We consider the following model variants:

\begin{itemize}
    \item \textbf{ViT + BiLSTM (FiLM Fusion)}: This variant uses only the geometric encoder (ViT) and the temporal encoder (BiLSTM), excluding the semantic land cover stream. The two feature types are fused using the FiLM mechanism.
    
    \item \textbf{HRNet + BiLSTM (FiLM Fusion)}: This variant uses a convolutional encoder (HRNet) to extract semantic information from land cover data, and a BiLSTM for meteorological encoding. The geometric branch is removed. Fusion is performed using FiLM.
    
    \item \textbf{ViT + HRNet + BiLSTM (Concat Fusion)}: This configuration includes both spatial encoders (ViT for nDSM and HRNet for land cover) along with the BiLSTM, but replaces the FiLM-based dynamic fusion with simple concatenation of the spatial features followed by joint processing.
    
    \item \textbf{GSM-UTCI (Ours)}: Our proposed full model combines all three modalities, geometric, semantic, and meteorological, using FiLM-based dynamic conditioning of both spatial encoders.
\end{itemize}

This set of experiments allows us to isolate the contribution of each encoder stream and compare fusion strategies, in order to validate the importance of multimodal inputs and cross-modal conditioning in urban heat stress prediction.

\subsection{Model implementation and validation}

\subsubsection{Model implementation}

The GSM-UTCI model was implemented using the PyTorch deep learning framework and trained on a high-performance computing server equipped with two NVIDIA RTX A6000 GPUs (48 GB each) and dual Intel Xeon Gold 6258R CPUs (2.70 GHz, 112 logical cores). Prior to training, all input data were normalized to ensure numerical stability. The nDSM and meteorological variables were standardized using z-score normalization:
\[
\hat{x} = \frac{x - \mu}{\sigma}
\]
where $x$ is the raw input value, and $\mu$ and $\sigma$ represent the mean and standard deviation computed from the training dataset. Categorical land cover values, ranging from 0 to 6, were scaled to the $[0, 1]$ interval by dividing by 6.

The model was trained using the AdamW optimizer with a learning rate of $1 \times 10^{-3}$ and weight decay of $1 \times 10^{-4}$, using a batch size of 24. The input data were processed as image tiles of size $512 \times 512$ pixels, matching the input resolution of each encoder. The training objective was to minimize Mean Squared Error (MSE) loss:
\[
\mathcal{L}_{\text{MSE}} = \frac{1}{n} \sum_{i=1}^n (y_i - \hat{y}_i)^2
\]
where $y_i$ and $\hat{y}_i$ represent the ground truth and predicted UTCI values, respectively. The model was trained for 1000 epochs, which proved sufficient for convergence and performance stability.

For model initialization, the geometric encoder adopted a \texttt{vit\_tiny\_patch16\_224} backbone, and the semantic encoder used \texttt{hrnet\_w18}, both pretrained on ImageNet. The full dataset consisted of 12,642 image tiles was randomly split into 70\% for training and 30\% for testing, ensuring a balanced distribution of geographic and climatic diversity across samples.

\subsubsection{Model evaluation metrics}

Model performance was evaluated using four common regression metrics: Mean Absolute Error (MAE), Mean Squared Error (MSE), Mean Absolute Percentage Error (MAPE), and the Coefficient of Determination ($R^2$). These metrics are defined as:

\begin{itemize}
    \item \textbf{Mean Absolute Error (MAE):}
    \[
    \text{MAE} = \frac{1}{n} \sum_{i=1}^{n} \left| y_i - \hat{y}_i \right|
    \]

    \item \textbf{Mean Squared Error (MSE):}
    \[
    \text{MSE} = \frac{1}{n} \sum_{i=1}^{n} (y_i - \hat{y}_i)^2
    \]

    \item \textbf{Mean Absolute Percentage Error (MAPE):}
    \[
    \text{MAPE} = \frac{100\%}{n} \sum_{i=1}^{n} \left| \frac{y_i - \hat{y}_i}{y_i} \right|
    \]

    \item \textbf{Coefficient of Determination ($R^2$):}
    \[
    R^2 = 1 - \frac{\sum_{i=1}^{n} (y_i - \hat{y}_i)^2}{\sum_{i=1}^{n} (y_i - \bar{y})^2}
    \]
    where $\bar{y}$ is the mean of the ground truth values.
\end{itemize}

Model outputs were compared against the SOLWEIG-generated UTCI maps as reference targets. All metrics were computed at the tile level and averaged across the validation dataset to assess generalization performance.

\subsection{Systematic land cover simulation analysis}

To evaluate the individual thermal contributions of different urban surface types, we conducted a systematic land cover simulation analysis using the GSM-UTCI model. The objective of this analysis is to quantify how various dominant land cover classes, such as buildings, impervious surfaces, bare earth, and vegetated areas, affect spatial patterns of outdoor heat stress across the city. By isolating the influence of each surface type within the predictive framework, we aim to generate transferable insights that can inform evidence-based landscape planning and climate adaptation strategies.

This experiment involves generating a series of counterfactual land cover maps in which each major surface class is systematically replaced—one at a time—with a fixed reference type, namely tree canopy. This reference was chosen to represent a realistic and widely promoted greening intervention, given the well-established cooling benefits of urban trees through shading and evapotranspiration. For each simulation, the meteorological conditions are held constant to ensure that observed differences in predicted UTCI can be attributed specifically to the altered land cover and structural form.

Because the nDSM plays a key role in determining radiation exchange and shading, it is also modified during the substitution process. When a land cover type is replaced with tree canopy, the corresponding nDSM values are reassigned using the average height of tree canopy in the same tile, computed from the original data. If no local tree canopy exists within a given tile, the city-wide mean tree height is used instead. This approach ensures spatial consistency while maintaining realistic assumptions about the three-dimensional structure of the landscape under the greening scenario.

A baseline UTCI map is first generated using the original land cover input. Then, for each substitution scenario, all pixels labeled with a target land cover class (e.g., impervious surfaces or bare earth) are reassigned to tree canopy in the input raster. These modified nDSM and land cover maps are then fed into the GSM-UTCI model to predict a new UTCI distribution under the hypothetical landscape condition.

The resulting UTCI maps allow us to calculate the change in thermal exposure ($\Delta$UTCI) associated with each substitution scenario at both the pixel and city-wide level. By comparing the spatial distribution and magnitude of temperature reductions, we are able to rank land cover types by their contribution to urban heat retention or mitigation. This type of structured simulation provides a rigorous, spatially explicit method for assessing the thermal benefits of land cover transformation strategies and can directly inform green infrastructure planning, zoning updates, and urban forestry investments.

\section{Results}
\label{as}

\subsection{Spatial-temporal distribution and patterns of UTCI}

\begin{figure}[H]
    \centering
    \includegraphics[width=\linewidth]{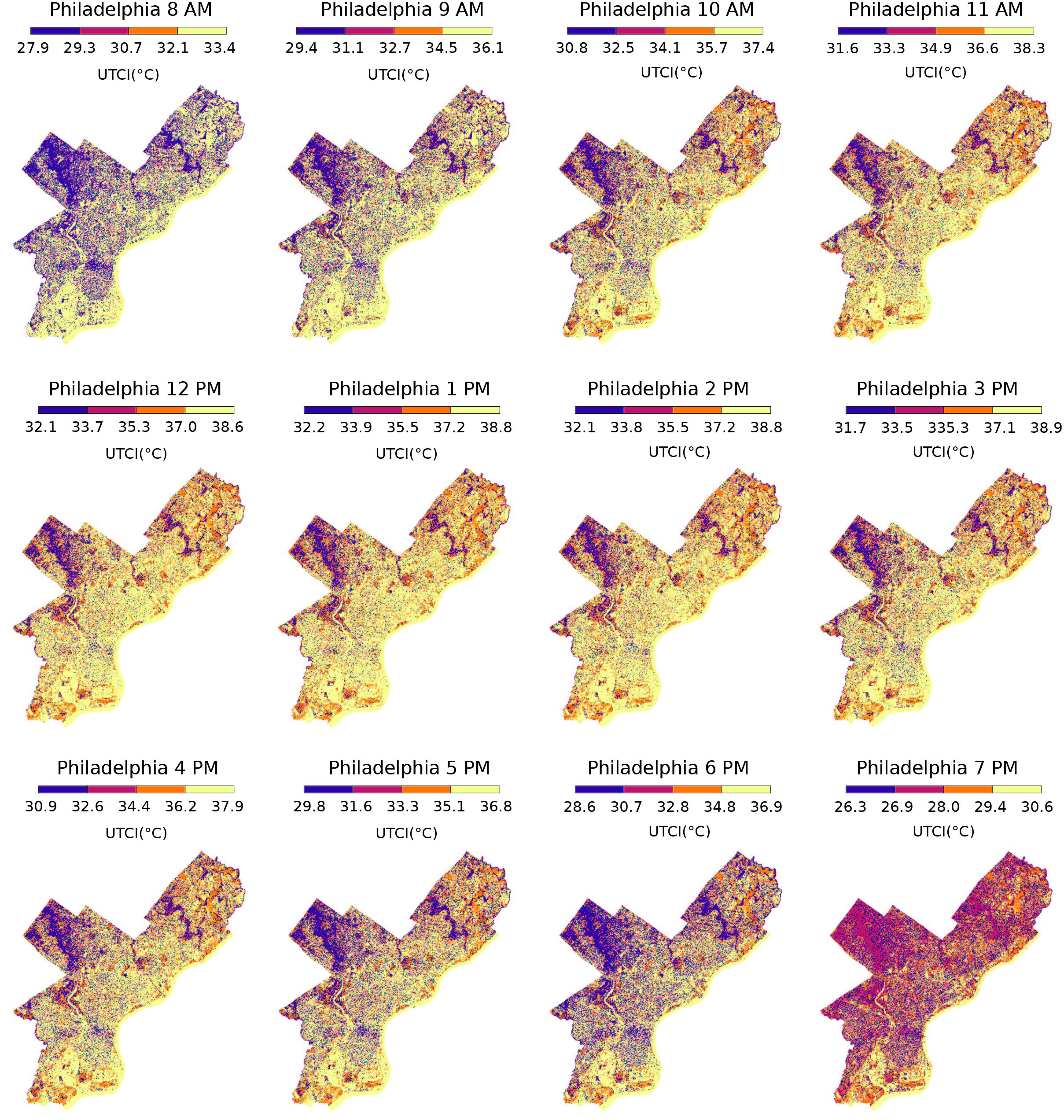}
    \caption{Hourly UTCI maps for Philadelphia from 8:00 a.m. to 7:00 p.m. based on SOLWEIG simulation. Higher UTCI values (red-yellow) indicate greater heat stress. Tree-covered and open green spaces consistently show lower UTCI levels compared to impervious built-up areas.}
    \label{fig:utci_hourly}
\end{figure}

\myfigref{fig:utci_hourly} presents the spatial-temporal distribution of UTCI values in Philadelphia over August. Eleven hourly maps, from 8 a.m. to 7 p.m., are shown at 1-meter resolution to capture the fine-scale variation of outdoor heat stress across the urban landscape. 
The results reveal a clear diurnal pattern of heat buildup and dissipation, with UTCI values rising steadily from morning to early afternoon, peaking between 1 and 3 p.m., and gradually decreasing in the late afternoon. During early morning hours (8 – 9 a.m.), UTCI values are generally below $32^\circ$C, indicating moderate thermal stress conditions in most areas. However, by midday (12 – 2 p.m.), large portions of the city experience UTCI levels exceeding $35^\circ$C, with localized hotspots surpassing $38^\circ$C, particularly in open spaces with minimal shading or vegetative cover.

Spatially, the highest UTCI values are generally observed in impervious vacant lands with limited shading or vegetation. In contrast, several core urban districts exhibit lower UTCI levels despite high development intensity. For example, Center City shows moderated thermal stress, likely due to dense high-rise structures that provide substantial shading during peak sun hours. Similarly, University City displays relatively lower UTCI values, benefiting from proximity to the riverfront and higher tree canopy coverage associated with institutional campuses. These spatial patterns highlight the complex interplay between urban form, vegetation, and solar geometry, particularly during peak heat periods when shading and evapotranspiration are most effective in mitigating outdoor thermal stress.

\begin{figure}[H]
    \centering
    \includegraphics[width=0.65\linewidth]{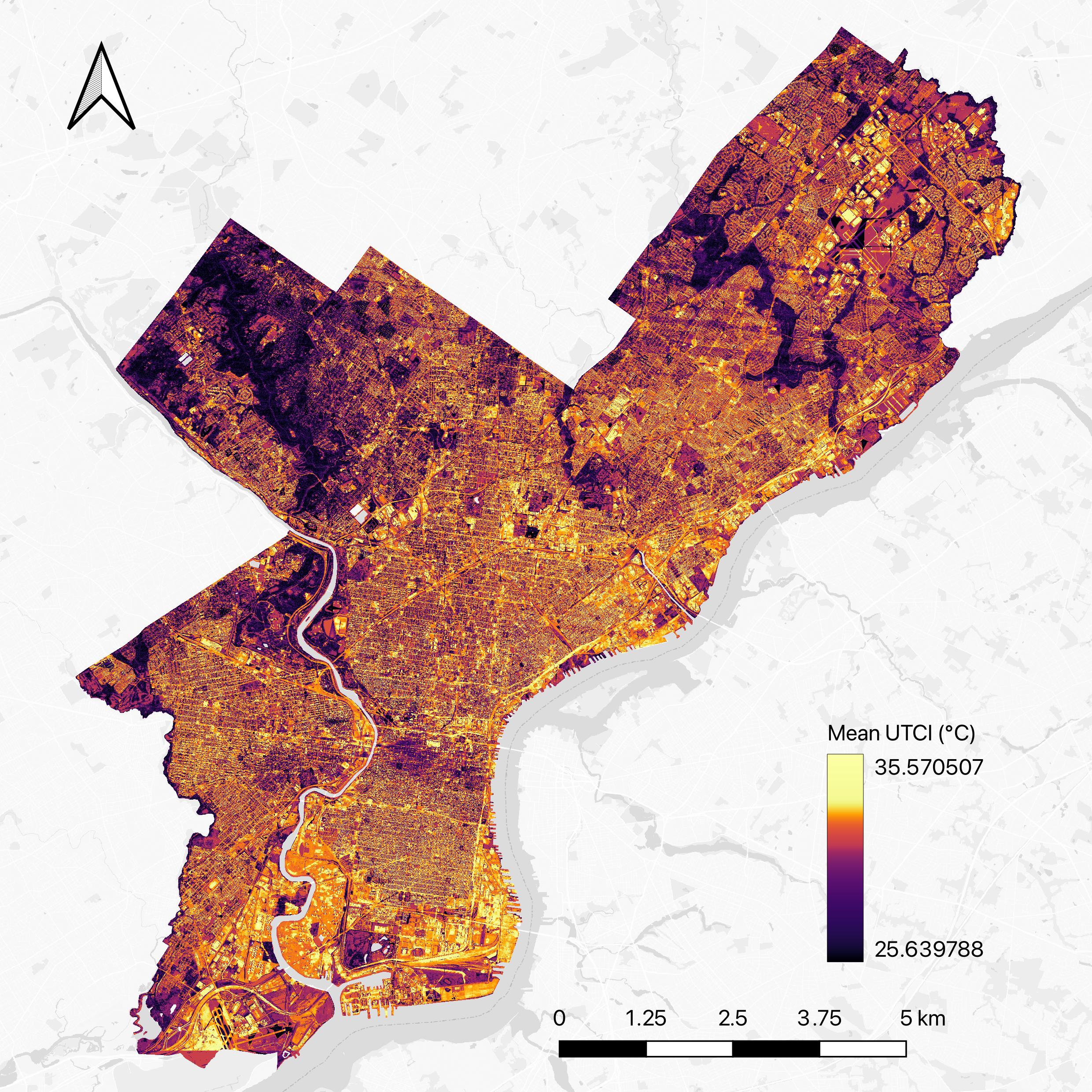}
    \caption{Average UTCI map for Philadelphia in Auguest, 2020.}
    \label{fig:mean_utci}
\end{figure}

In the late afternoon and early evening (4 – 7 p.m.), UTCI values begin to decline, though residual heat remains elevated in thermally massive areas such as asphalt-dominated streets and rooftops. By 7 p.m., UTCI values in most vegetated areas have returned to below $30^\circ$C, while built-up zones still exhibit delayed cooling. 
Overall, the SOLWEIG-derived hourly UTCI maps demonstrate strong spatial-temporal heterogeneity in urban heat exposure, emphasizing the influence of land surface characteristics and urban morphology on thermal comfort conditions throughout the day. Based on these hourly outputs, we compute the daytime average UTCI from 8 a.m. to 7 p.m., which serves as the target variable for model training and evaluation in this study (\myfigref{fig:mean_utci}).

\subsection{Ablation studies and model comparison}

\begin{table}[H]
\centering
\caption{Ablation and model comparison for UTCI prediction across 12,642 validation tiles (512×512).}
\label{tab:model_performance}
\begin{tabular}{lccccc}
\toprule
\textbf{Model Variant} & \textbf{Params} & \textbf{MAE (°C)} & \textbf{MSE (°C$^2$)} & \textbf{MAPE (\%)} & \textbf{$R^2$} \\
\midrule
\textbf{\begin{tabular}[c]{@{}c@{}}A1: ViT + BiLSTM\\(FiLM Fusion)\end{tabular}} & 5,805,778  & 0.7394 & 1.2115 & 2.4650\% & 0.7690 \\
\textbf{\begin{tabular}[c]{@{}c@{}}A2: HRNet + BiLSTM\\(FiLM Fusion)\end{tabular}} & 11,099,238  & 0.4406 & 0.5285 & 1.4710\% & 0.8992 \\
\textbf{\begin{tabular}[c]{@{}c@{}}A3: ViT + HRNet + BiLSTM\\(Concat Fusion)\end{tabular}} & 16,718,839  & 0.4435 & 0.5035 & 1.4794\% & 0.9046 \\
\textbf{GSM-UTCI (Ours)} & 16,798,103   & \textbf{0.4130} & \textbf{0.4477} & \textbf{1.3750\%} & \textbf{0.9151} \\
\bottomrule
\end{tabular}
\end{table}

Table~\ref{tab:model_performance} presents the results of our ablation experiments and model comparisons on the validation dataset, consisting of 12,642 tiles at 512×512 resolution. We use four metrics to comprehensively assess prediction accuracy: MAE, MSE, MAPE, and the $R^2$. Our full model, GSM-UTCI, achieves the best performance across all metrics, with a MAE of 0.4130$^\circ$C, MSE of 0.4477 ($^\circ$C$^2$), MAPE of 1.3750\%, and $R^2$ of 0.9151. This confirms the effectiveness of incorporating both geometric and semantic spatial information, modulated by temporal weather dynamics through the FiLM fusion mechanism.

The A1 variant using only ViT and BiLSTM excludes semantic land cover information and performs significantly worse ($R^2 = 0.7690$, MAE = 0.7394), indicating that surface morphology alone is insufficient for accurate UTCI prediction. In contrast, the A2 variant using only the semantic encoder (HRNet) and BiLSTM achieves substantially better performance ($R^2 = 0.8992$, MAE = 0.4406), highlighting the dominant role of land cover characteristics in shaping urban heat stress. Nevertheless, both single-stream variants are outperformed by the full GSM-UTCI model, confirming the added value of integrating structural and semantic modalities via multimodal learning.

Furthermore, the A3 model variant, which fuses ViT and HRNet features via naive channel-wise concatenation, shows slightly improved accuracy ($R^2 = 0.9046$) over the HRNet-only model but still lags behind the proposed GSM-UTCI. This performance gap demonstrates the superiority of the FiLM-based fusion strategy, which allows meteorological conditions to dynamically modulate spatial features, leading to more context-sensitive and physically consistent predictions. 

\begin{figure}[H]
    \centering
    \includegraphics[width=\linewidth]{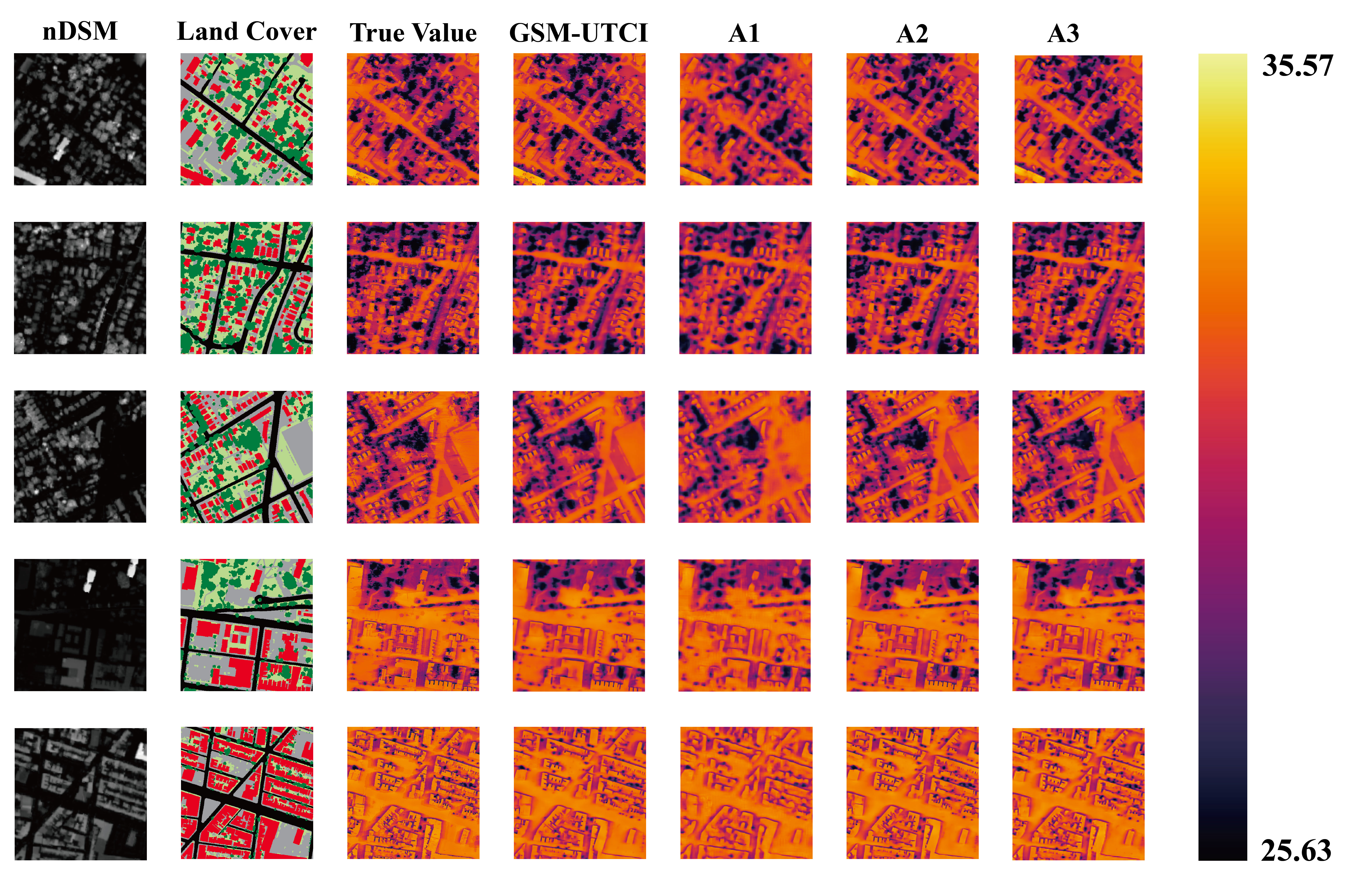}
    \caption{Visual comparison of UTCI predictions produced by the proposed GSM-UTCI model and three ablation baselines across five urban tiles. From left to right: input nDSM, land cover map, ground truth UTCI, GSM-UTCI prediction, and outputs from A1 (ViT + BiLSTM), A2 (HRNet + BiLSTM), and A3 (Concat Fusion). GSM-UTCI more accurately preserves shading and spatial heterogeneity than other variants, especially around vegetated and built-up transitions.}
    \label{fig:qualitative}
\end{figure}

\myfigref{fig:qualitative} presents a visual prediction comparison between the GSM-UTCI model and its ablated variants across five representative urban tiles. 
The GSM-UTCI model demonstrates a strong ability to preserve sharp thermal gradients and capture fine-scale features such as tree shadows and street-level shading. Boundaries between different surface types (e.g., vegetation, pavement, rooftops) are clearly defined, and areas with tall structures or dense canopy exhibit appropriate cooling effects.

In contrast, A1 (ViT + BiLSTM) shows significant spatial blurring and fails to delineate land surface boundaries, indicating the absence of semantic information severely hinders spatial accuracy. A2 (HRNet + BiLSTM) produces more structured predictions but lacks solar-induced heterogeneity, particularly in shaded zones, due to the exclusion of nDSM input. A3 (Concat Fusion) captures both morphology and semantics to some extent but struggles to represent cross-modal interactions accurately, resulting in flattened outputs and loss of local shading nuance. These differences highlight the necessity of both spatial modalities and the importance of context-aware fusion in urban heat modeling.

In addition to improved accuracy, GSM-UTCI demonstrates strong computational efficiency. It can generate citywide UTCI maps in approximately 5 minutes, which reduces runtime by orders of magnitude compared to traditional physical models such as SOLWEIG. This combination of high precision and rapid inference makes GSM-UTCI well-suited for large-scale planning applications.

\subsection{Systematic land cover simulation results}

Table~\ref{tab:substitution_summary} presents a quantitative summary of the thermal mitigation potential across three systematic land cover substitution scenarios: Bare Earth → Tree Canopy, Grass → Tree Canopy, and Impervious Surfaces → Tree Canopy. For each scenario, we calculate the affected area, average change in UTCI ($\Delta$UTCI), standard deviation (SD), post-substitution UTCI, and total aggregated thermal benefit measured in Kelvin square meters (K·m$^2$). Among the scenarios, converting impervious surfaces to tree canopy produced the highest total cooling potential (1,132.21M K·m$^2$), reflecting both a substantial average $\Delta$UTCI of --4.18\,$^\circ$C and a large spatial extent (270.66\,km$^2$). Although Bare Earth exhibits the strongest per-pixel cooling effect (--4.87\,$^\circ$C), its limited spatial coverage (23.15\,km$^2$) results in a lower overall impact. The Grass → Tree scenario offers moderate cooling benefits (--2.90\,$^\circ$C on average) over a larger area, underscoring the spatial trade-offs inherent in different greening strategies. These results demonstrate that land cover transformation toward increased tree canopy yields significant improvements in thermal comfort across varying urban surface types.

\begin{table}[H]
\centering
\caption{Summary statistics for land cover substitution scenarios.}
\label{tab:substitution_summary}
\begin{tabular}{lcccccc}
\toprule
\textbf{Scenario} &
\begin{tabular}[c]{@{}c@{}}\textbf{Area}\\ \textbf{(km$^2$)}\end{tabular} &
\begin{tabular}[c]{@{}c@{}}\textbf{Avg}\\ $\Delta$UTCI (°C)\end{tabular} &
\begin{tabular}[c]{@{}c@{}}\textbf{SD}\\ (°C)\end{tabular} &
\begin{tabular}[c]{@{}c@{}}\textbf{Post-}\\ \textbf{UTCI (°C)}\end{tabular} &
\begin{tabular}[c]{@{}c@{}}\textbf{Total}\\ $\Delta$UTCI\\ (K·m$^2$)\end{tabular} \\
\midrule
Bare Earth → Tree       & 23.15   & -4.87   & 1.32  & 27.41   & 112.83M \\
Grass → Tree      & 281.09     & -2.90     & 1.58   & 27.52     & 815.95M   \\
Impervious Surfaces → Tree   & 270.66     & -4.18     & 1.89   & 27.75     & 1,132.21M   \\
\bottomrule
\end{tabular}
\end{table}

\begin{figure}[H]
    \centering
    \includegraphics[width=\linewidth]{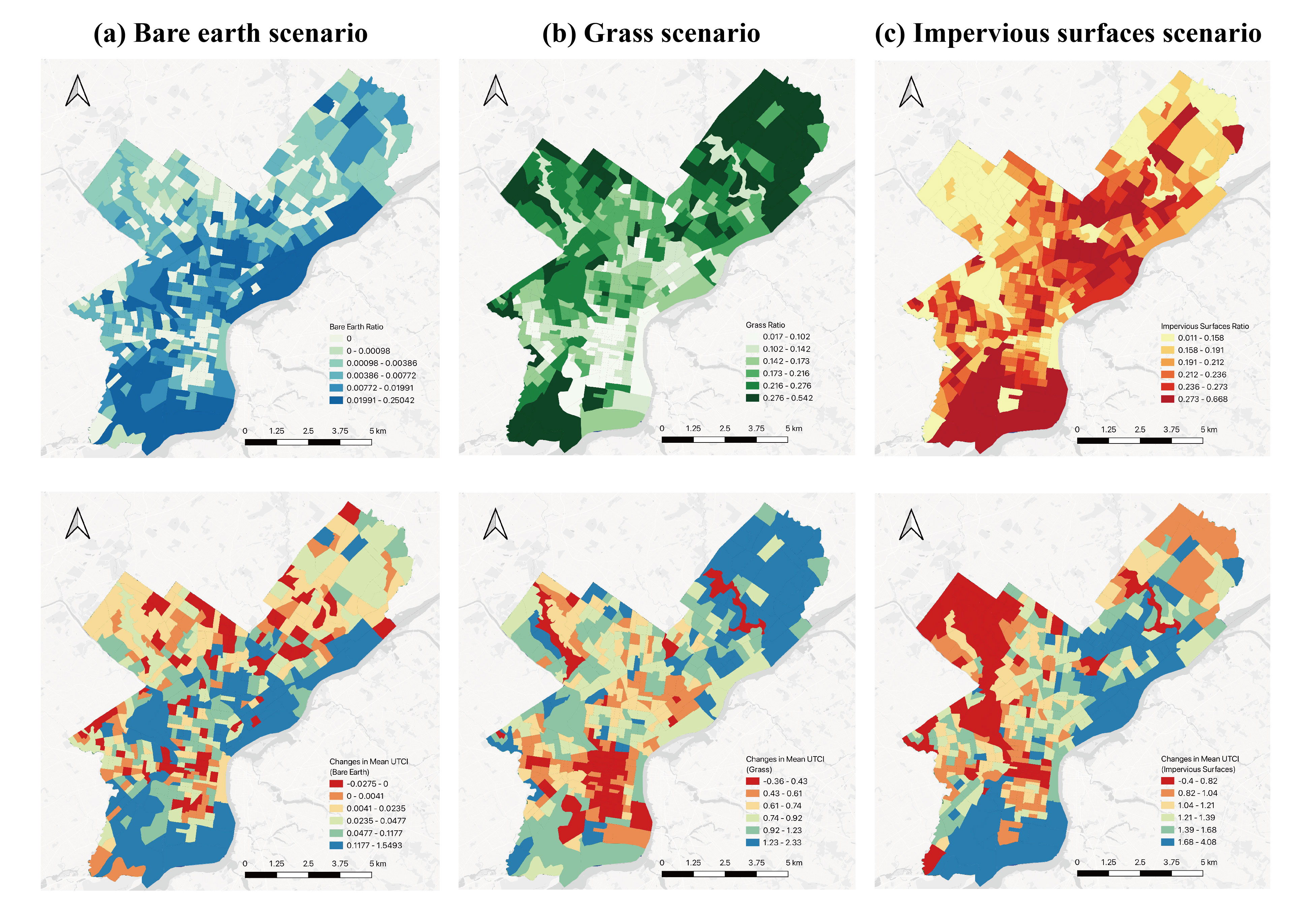}
\caption{Spatial distribution of land cover proportions (top row) and predicted change in mean UTCI (bottom row) at the census tract level for three substitution scenarios: (a) Bare Earth, (b) Grass, and (c) Impervious Surfaces to Tree Canopy.}
    \label{fig_SC}
\end{figure}

Figure~\ref{fig_SC} visualizes the spatial distribution of land cover ratios (top row) and the corresponding changes in mean UTCI (bottom row) at the census tract level for the three greening scenarios. Each pair of maps provides a complementary perspective: the upper panels illustrate the baseline proportion of target land cover classes, Bare Earth, Grass, and Impervious Surfaces, while the lower panels map the modeled $\Delta$UTCI resulting from converting these classes to tree canopy. In all three cases, the cooling benefits are spatially heterogeneous, with higher $\Delta$UTCI observed in tracts with greater initial coverage of heat-intensive surfaces. For instance, in the Bare Earth scenario, tracts with even small proportions of exposed soil show substantial localized reductions in thermal stress. In the Impervious Surfaces scenario, the most intense cooling effects are concentrated in the central and southern tracts, where dense built environments and minimal vegetation dominate. The visual correspondence between land cover abundance and thermal reduction supports the interpretation that landscape structure strongly mediates the effectiveness of urban greening interventions.

\section{Discussion}
\label{d}

\subsection{Modeling accuracy and reliability}

The GSM-UTCI model represents a significant advancement in the modeling of urban heat stress by transferring traditional physical simulation frameworks into a scalable deep learning paradigm. Unlike physics-based models such as SOLWEIG, which require detailed radiative transfer calculations and often demand hours of computation per city-wide simulation, our approach achieves comparable predictive accuracy at substantially reduced computational cost. Specifically, GSM-UTCI can produce high-resolution UTCI maps for an entire city (e.g., Philadelphia) in under five minutes, enabling efficient scenario testing and large-scale planning support. 
Empirical validation demonstrates strong predictive performance, with a coefficient of determination ($R^2$) of 0.9151, a mean absolute error (MAE) of 0.41\,$^\circ$C, and a mean absolute percentage error (MAPE) below 2\%. These results confirm the model's capability to generalize across diverse urban morphologies and meteorological conditions.

\subsection{Landscape and urban planning implications}

The results of this study provide actionable insights for landscape and urban planning strategies aimed at mitigating outdoor thermal stress. First, the simulation experiments demonstrate that targeted land cover transformations, particularly converting impervious surfaces and bare earth to tree canopy, can substantially reduce UTCI at the neighborhood scale \citep{ziter2019scale, yi2025assessing, nowak2012tree}. This highlights the importance of integrating urban forestry and surface greening as core components of heat resilience planning.

Second, the tract-level bivariate analysis reveals strong spatial heterogeneity in both existing surface composition and cooling potential. This suggests that universal greening policies may be inefficient or inequitable, and instead supports the use of data-informed, spatially targeted interventions. High-priority zones include areas with both high impervious coverage and high UTCI, which were shown to benefit most from tree planting interventions.

Finally, the high-resolution, scenario-driven nature of GSM-UTCI allows planners to evaluate not only where to intervene, but also how specific land cover changes may impact thermal comfort. This capability supports the development of precision adaptation strategies, such as evaluating trade-offs between vegetative types, simulating incremental greening scenarios, or integrating heat mitigation into zoning and land-use policy. As cities seek to address both climate adaptation and environmental equity, the model provides a scalable, interpretable, and practical tool for aligning design decisions with microclimatic performance outcomes.

\subsection{Limitations and future directions}

While the GSM-UTCI model demonstrates strong performance and operational efficiency, several limitations should be acknowledged. First, the model relies on spatially static nDSM and land cover inputs, which do not capture dynamic shading or diurnal morphological changes. Second, although the current architecture implicitly captures heat-retaining effects of built surfaces, it does not explicitly model radiative mechanisms such as shadowing, albedo variation, or longwave radiation exchange. These omissions may lead to local prediction errors in areas with complex building forms or rapidly changing insolation conditions. Lastly, this study focused on a single city (Philadelphia); while the model is designed for generalizability, its applicability across cities with different climatic zones and urban typologies has not yet been empirically validated.

Looking forward, several promising directions could extend the capabilities and impact of this framework. From a data perspective, incorporating additional environmental factors, such as dynamic solar shadows, surface albedo maps \citep{yi2025sub}, and real-time radiation datasets, may improve the accuracy of fine-scale thermal predictions. From a spatio-temporal perspective, extending the model to predict hourly or seasonal UTCI sequences across cities in different climate zones would enhance its value for regional climate adaptation planning. Furthermore, the GSM-UTCI framework could be adapted for prescriptive applications: for example, by systematically modifying land cover compositions or tree planting distributions, planners could use the model to simulate and optimize greening strategies, quantify marginal cooling effects, and design equitable interventions tailored to local needs.


\section{Conclusion}
\label{c}

This study introduces GSM-UTCI, a multimodal deep learning framework for predicting and simulating human-perceived urban heat stress at hyperlocal resolution. By integrating surface morphology, land cover, and temporally dynamic meteorological data through a feature-wise linear modulation (FiLM) mechanism, the model effectively replicates SOLWEIG-derived UTCI patterns while significantly reducing computational time. GSM-UTCI achieves an $R^2$ of 0.9151 and mean absolute error of 0.41\,$^\circ$C across a diverse urban landscape, with the ability to generate city-wide UTCI maps at 1-meter resolution in under five minutes.

Beyond prediction, the framework supports scenario-based simulations of landscape transformation, allowing planners to evaluate how specific land cover interventions, such as increasing tree canopy, can mitigate thermal stress at the neighborhood scale. Our simulation results in Philadelphia demonstrate that converting impervious surfaces and bare earth to vegetated cover yields substantial cooling benefits, especially in high-density and low-canopy tracts.

In conclusion, these findings highlight the potential of GSM-UTCI to serve as a scalable and practical decision support tool for climate-responsive urban design and planning. Future research could expand this framework to multi-city and multi-climate contexts, incorporate additional environmental factors such as shadow dynamics and surface albedo, and apply the model to optimize spatial configurations of greening strategies. By bridging the gap between environmental simulation and actionable planning, GSM-UTCI contributes a timely tool for building heat-resilient cities.








\bibliographystyle{model5-names}
\bibliography{main}
\end{document}